\documentclass[default,iicol,sn-mathphys]{sn-jnl}

\jyear{2024}%

\theoremstyle{thmstyleone}%
\theoremstyle{thmstyletwo}%

\theoremstyle{thmstylethree}%

\usepackage{diagbox}
\usepackage{ulem}
\usepackage{soul}

\begin{document}

\title[Article Title]{When the Inference Meets the Explicitness or Why Multimodality Can Make Us Forget About the Perfect Predictor}


\author*[1,2]{\fnm{J. E.} \sur{Dom\'{i}nguez-Vidal}}\email{jdominguez@iri.upc.edu}

\author[1,2]{\fnm{Alberto} \sur{Sanfeliu}}\email{alberto.sanfeliu@upc.edu}

\affil*[1]{\orgname{Institut de Robòtica i Informàtica Industrial (CSIC-UPC)},
\orgaddress{\street{Llorens i Artigas 4-6}, \city{Barcelona}, \postcode{08028}, \country{Spain}}}

\affil[2]{\orgname{Universitat Politècnica de Catalunya~-~BarcelonaTech (UPC)}, 
\orgaddress{\street{Jordi Girona, 31}, \city{Barcelona}, \postcode{08034}, \country{Spain}}}



\abstract{Although in the literature it is common to find predictors and inference systems that try to predict human intentions, the uncertainty of these models due to the randomness of human behavior has led some authors to start advocating the use of communication systems that explicitly elicit human intention. In this work, it is analyzed the use of four different communication systems with a human-robot collaborative object transportation task as experimental testbed: two intention predictors (one based on force prediction and another with an enhanced velocity prediction algorithm) and two explicit communication methods (a button interface and a voice-command recognition system). These systems were integrated into IVO, a custom mobile social robot equipped with force sensor to detect the force exchange between both agents and LiDAR to detect the environment. The collaborative task required transporting an object over a $5$–$7$ meter distance with obstacles in the middle, demanding rapid decisions and precise physical coordination. $75$ volunteers perform a total of $255$ executions divided into three groups, testing inference systems in the first round, communication systems in the second, and the combined strategies in the third. The results show that, 1) once sufficient performance is achieved, the human no longer notices and positively assesses technical improvements; 2) the human prefers systems that are more natural to them even though they have higher failure rates; and 3) the preferred option is the right combination of both systems.}




\maketitle

\section{Introduction}

Originally, robotics began by performing small, routine and repetitive tasks to relieve humans of this burden. As perception systems improved, robots were able to better perceive and understand the environment around them allowing them to perform more complex tasks by being able to make the right decision when navigating an urban environment~\cite{goldhoorn2017} or choosing the right tool~\cite{saito2021}. However, when these machines started to interact with humans, perceiving the environment was no longer enough and we started to need to understand the human's intention as well since their behavior was too uncertain~\cite{dragan2017, choudhury2019}.

This is when we began to develop predictors to allow the robot to infer human intent with increasingly higher success rates~\cite{ordonez2016, schydlo2018}. While their non-negligible error rates are often blamed on limitations in computational capacity or a lack of sufficient data for training, the fact that humans can model the same information we perceive in multiple ways~\cite{johnson1989} making it so that two humans can represent the same environment differently begs the question of whether we will ever have a perfect predictor.

This very question has caused some authors~\cite{gildert2018, lee2018, dar2020, Inferencia_ROMAN2023} to begin to consider it necessary to combine inference engines with communication systems that make it possible to obtain the human's intention explicitly. In this way, they hope to achieve robots that do not just look like humans, but act like humans: they ask questions and request information when the one they have does not allow them to make decisions with sufficient confidence.

Thus, this work arises as a continuation of our previous work~\cite{Inferencia_ROMAN2023} where in a collaborative transportation task we first confronted on the one hand an inference system to obtain the human's intention from the instantaneous force they are exerting against, on the other hand, a button-based communication system to allow the human to express their intention explicitly. That work proved that allowing humans to directly express themselves can achieve the same improvement in multiple aspects of an effective Human-Robot Interaction (HRI) as using an intention predictor. However, it also left unanswered questions, such as what happens if both systems are combined, and postulated that we should not continue looking for technical improvements in the predictor used, since these may go unnoticed by the human, but that we should pivot towards methods that seek to improve human-robot communication, although without demonstrating these assertions. This work comes to test those considerations analysing human preferences at the time of working with a robot in a task with fast physical contact and to answer that question through three rounds of experiments with their respective user studies.

Thus, our contributions would be as follows. In the first round of experiments, we compare two predictors with different success rates and find that, once the failure rate is reduced to an acceptable value, the human no longer perceives any improvement. In the second round of experiments, we compare two direct communication systems and find that the human prefers the one that is more natural even though it is technically inferior in terms of response delay and failure rate. Finally, we compare the system preferred by the human in each of the first two rounds of experiments as well as the combination of both to verify that this combination is the most accepted by the human as it offers more freedom to collaborate with the robot, being this our third contribution.

In the remainder of the document, we present the related work in Section~\ref{sec:related_work}. Section~\ref{sec:collaborative_transportation} includes an explanation of the task selected for this study, the collaborative transport of objects, as well as all the relevant details of all the systems employed in this work. Section~\ref{sec:evaluation} presents the hypotheses to be tested, the setup and methodology employed as well as the distribution of participants who performed the experiments. Section~\ref{sec:results} presents the results obtained in each round of experiments and Section~\ref{sec:discussion} shows a short discussion of these results. Finally, Section~\ref{sec:conclusions} contains the conclusions.

\section{Related Work}\label{sec:related_work}

Of the two strategies discussed in the previous Section to know human's intention, inference engines and direct communication systems, the former is relatively abundant in the literature~\cite{luo2019, jain2018, maroger2021, thobbi2011}. Most of these works use different architectures based on Gaussian Mixture Models (GMM) or some kind of Artificial Neural Network (ANN) to obtain a prediction of the human's intention whether this is the trajectory they are going to follow, the movement they are going to make with their hand or the next object they are going to pick up. Applied to collaborative transportation tasks or, in general, tasks with physical contact, it has been common to use control techniques to make the robot adapt to the human's wishes using both impedance and admittance controllers~\cite{agravante2014, bussy2012, tarbouriech2019, yu2021, li2017}. More recent work, on the other hand, has to include some kind of predictor, either of the desired trajectory for the object~\cite{alevizos2020} or of its velocity profile~\cite{alyacoub2021} or even of the force that the human is going to exert on the object in the short term~\cite{Predictor_IROS2023}. None of these works contemplate the possibility of the human communicating with the robot to reduce uncertainty or resolve any problems generated by an error in the robot's inference.

This second strategy, a direct communication system, is less common. In~\cite{ROMAN2021} the authors design a smartphone application with which human and robot can communicate bidirectionally over long distances. This app is applied in~\cite{Dalmasso2023} to collaborative search. Thanks to it, the robot can know both the route that the human is following and the one they want to follow even in urban scenarios with multiple occlusions that do not allow the robot to track the human. This makes it possible to minimize the overlapping of the areas explored by each agent. Another example where the human is allowed to explicitly communicate with the robot is~\cite{mullen2021}. In this, the robot infers the goal of the task that the human wants the robot to perform, but before executing it, it asks for confirmation from the human through an augmented reality (AR) based system. In~\cite{gildert2022} both strategies are combined to improve object manipulation between two robots. To this end, both robots communicate their plans both implicitly through the force they exert on the object and explicitly by exchanging wireless messages. Finally,~\cite{lorentz2023} allows the human to use a combination of gestures and voice commands to tell the robot which object to pick up and where to take it. 

Regarding voice commands recognition and applied to robotics, although there are old works that tried to allow the human to transmit simple movement commands to a mobile robot~\cite{rogalla2002, lv2008}, it was not until the proliferation of Artificial Neural Networks (ANN) and the emergence of large datasets containing lists of typical commands~\cite{warden2018} that satisfactory results began to be achieved by detecting finite lists of commands~\cite{majumdar2020, kim2021}. With the exception of~\cite{mullen2021}, none of the above works compares their communication system with another one. Moreover, all of them assume that the human is willing to use their system. In this article, we  do not take that for granted and perform both the comparison of multiple systems and that verification. Additionally, and to the best of our knowledge, this is the first work that tries to combine both strategies applied to this use case.



\section{Collaborative Object Transportation}\label{sec:collaborative_transportation}

\begin{figure}[t]
    \centering
    \includegraphics[width=0.65\linewidth]{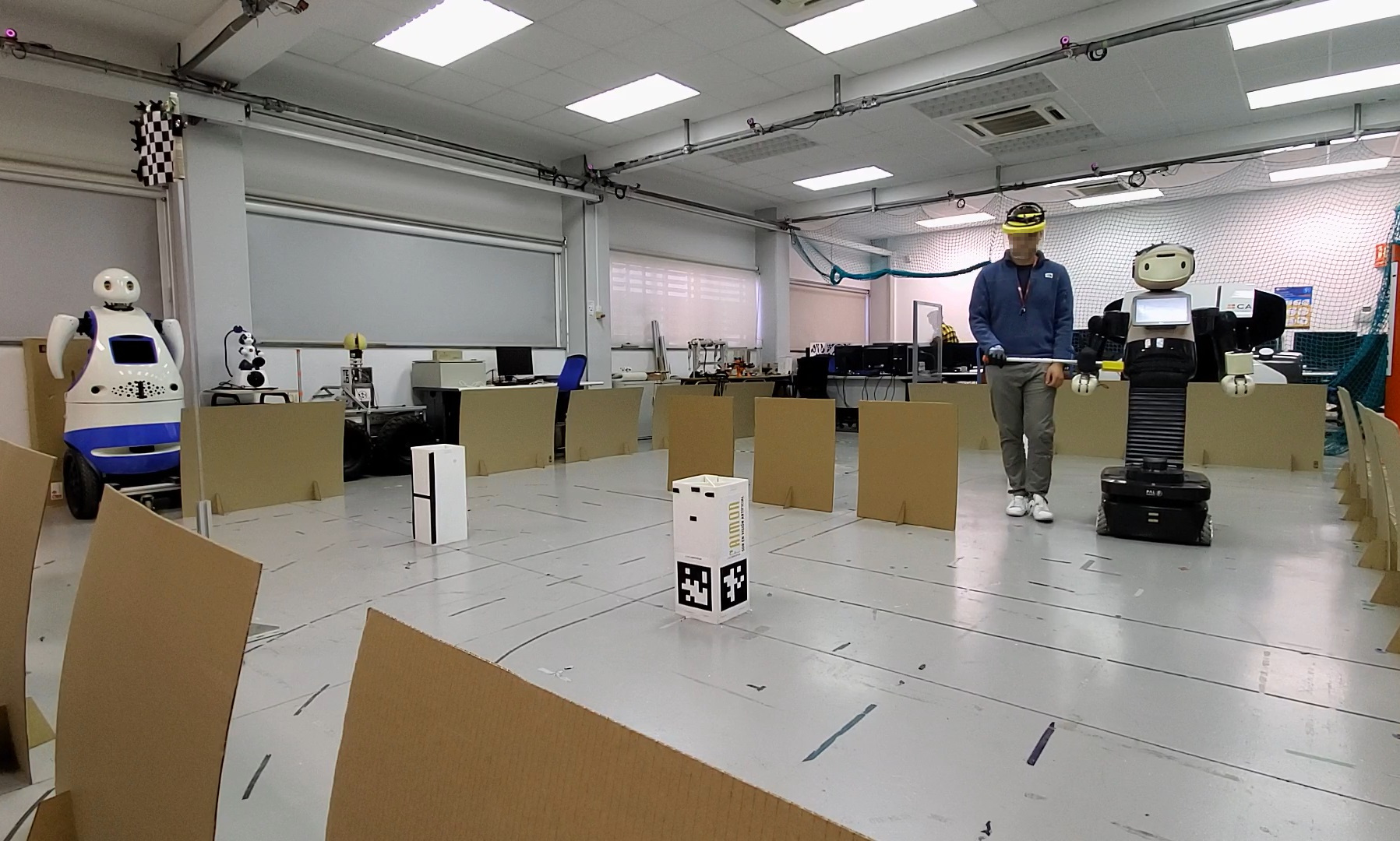}
    \includegraphics[width=0.33\linewidth]{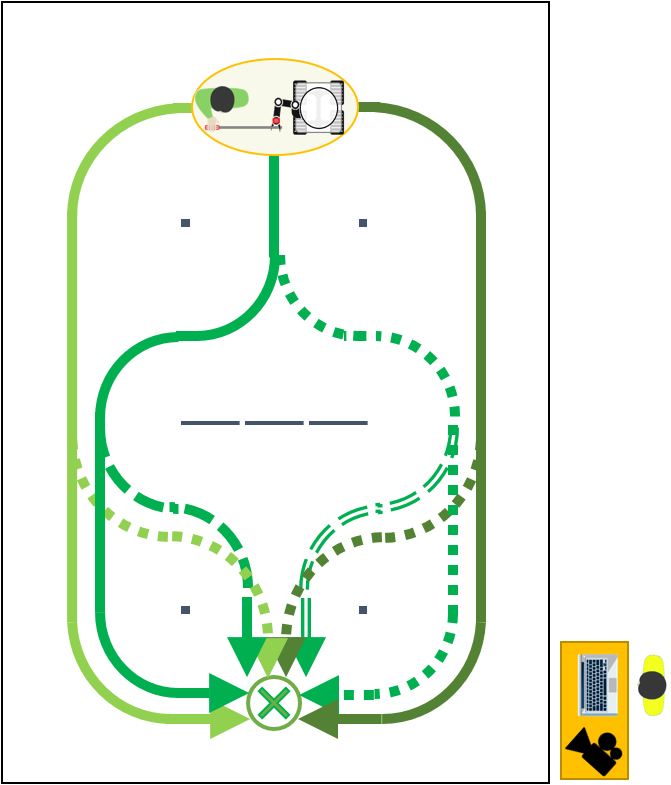}
    \includegraphics[width=0.62\linewidth]{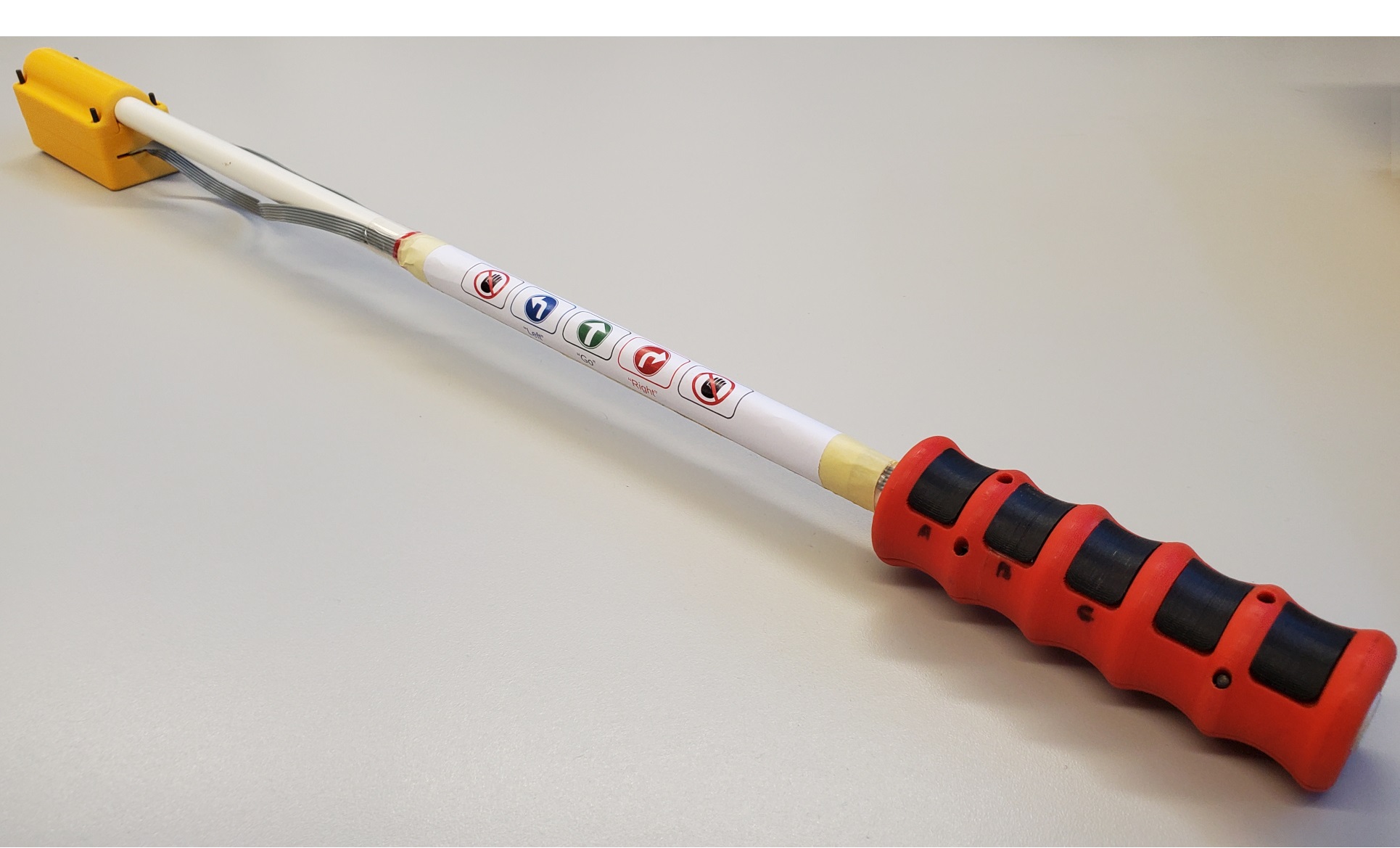}
    \includegraphics[width=0.36\linewidth]{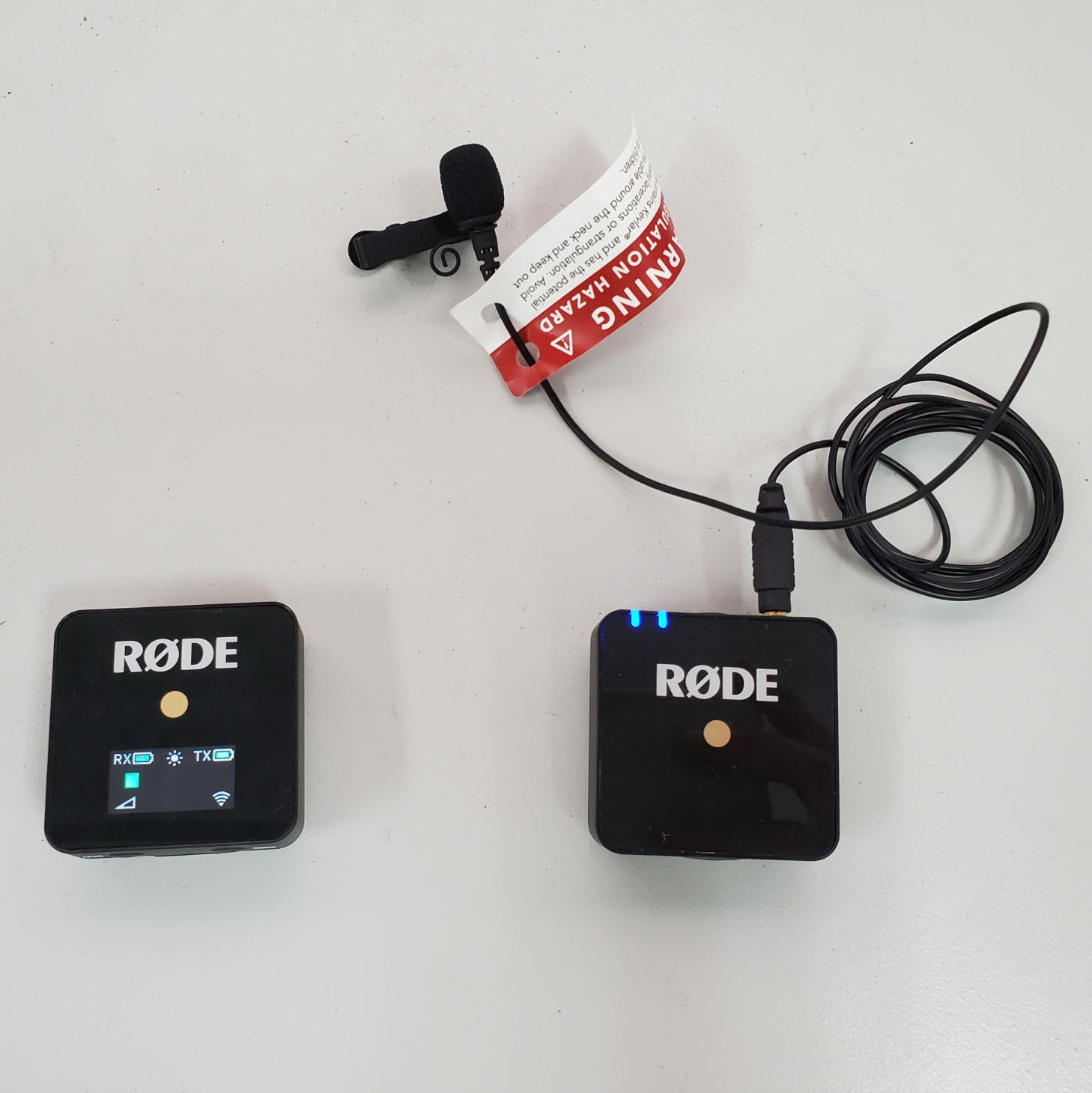}
    \caption{{\bf Experiments setup.} {\it Top Left} - Human-robot pair collaboratively transporting an aluminium bar. Goal marked with a chequered flag. {\it Top Right} - Scheme of the designed setup. At least eight routes to the goal. Control desk on the right with researcher managing the experiment and camera recording the point of view on the left. {\it Bottom Left} - Handle for better ergonomics with meaning of each button next to it. Only three buttons are used to tell the robot which route the human wants. {\it Bottom Right} - RODE Wireless GO microphone used for voice command recognition.}
    \label{fig:human-robot_pair_experiment}
\end{figure}

Collaborative object transportation is a task in which a human and a robot collaboratively transport an object from a starting point to a destination point that may or may not be predefined in advance. It is therefore a task that is performed in close proximity and in which there is an exchange of physical forces between the two agents. These two characteristics mean that the robot's movements cannot be abrupt, so as not to harm the human next to it, and that the robot's response speed must be high in order to be able to adapt to the rapid changes that the human may make.

Additionally, this task usually involves moving an object at a short distance using only a robotic arm. In this work, we will transport an object $5$-$7$~$m$ through a scenario with multiple obstacles so that there are multiple routes to the predefined goal or even change the route on the fly. This implies that the robot must move its platform through the scenario adapting to the route desired by the human and that the human must make multiple decisions along this route (see Fig.~\ref{fig:human-robot_pair_experiment}~-~{\it Top}).

To enable the robot to perform this task, we start from the implementation in~\cite{PIA_Journal}. In this, taking advantage of the fact that this is a task in which the exchange of information is mainly done through forces, the environment is modeled by means of virtual forces: a repulsive force for each obstacle detected by the robot and an attractive force for the partial goals that the human-robot pair should follow to reach the place where to leave the object. These partial goals are obtained from the waypoints generated by a global planner. The force resulting from modeling the environment is then combined with the force exerted by the human on one end of the transported object and measured by a force sensor on the robot's wrist to which the other end of the object, in this case an aluminum bar, is attached. This combined force is sent to a controller that generates the robot's speed commands.

Additionally, this system is conditioned using the human intention. For this purpose, we start from the distinction between implicit and explicit intention shown in~\cite{PIA_HRI2023}. Thus, we consider as implicit intention that which can be inferred from the actions of the other agent and as explicit intention that which is expressed using a direct communication channel and a code known and agreed upon by both agents.

\subsection{Inference of Implicit Intention}

Two similar inference systems, representative of the trend in the literature towards using predictors, will be used to obtain the implicit intention of the human. First, a force predictor which, from the previous values of the force exerted by the human, the velocity of the human-robot pair, the representative force of the environment and the robot's LiDAR readings, generates a prediction of the force that the human will exert during the next $1$~$s$. This prediction can be processed to obtain an estimate of the route the human wishes to follow in the short term, and with this, condition the robot's planner to match their intention.

This force predictor used that way has a fundamental shortcoming and that is that when predicting the route it only takes into account the contribution of the human through their force and not the contribution performed by the robot, e.g., avoiding getting too close to an obstacle or damping rapid changes in the force exerted by the human. This is the reason why in our previous work~\cite{Inferencia_ROMAN2023} we suggested that the predictor used had room for improvement.

To solve this, we develop a second predictor that, in addition to the force to be exerted by the human, predicts the velocity of the human-robot pair during the next $1$~$s$. This second prediction can be directly integrated to obtain an estimate of the short-term desired route with which to also condition the robot's planner. While the technical details of this second predictor are outside the scope of this paper~\cite{DominguezVidal2024_ICRA}, it allows to reduce the L2 error made in estimating the trajectory at $1$~$s$ from $0.199$~$m$ with the force predictor to $0.138$~$m$ with this velocity predictor.

\subsection{Direct Communication of Explicit Intention}

To obtain the explicit intention of the human we also used two systems. First, the same system with three buttons on the object's handle used in~\cite{Inferencia_ROMAN2023}. By pressing each of them, the human can tell the robot whether to go straight ahead, turn left or turn right. Information that the robot uses to condition its planer at the next intersection. The second system implemented is a voice command recognition model. With this, the human can verbally tell the robot their intention using the 'Go', 'Left' and 'Right' commands. These, generate the same conditioning as with the previous buttons (see Fig.~\ref{fig:human-robot_pair_experiment}~-~{\it Bottom}).

The system with buttons is infallible in that the only possible error is that the human presses the wrong button and its processing delay is negligible. Meanwhile, voice command recognition can make mistakes (hit rate: 94.75\%). Moreover, if a Bluetooth microphone is used as in our case, the delay between the human saying a command and it taking effect can go from $0.2$ to $1$~$s$ if the RF cell is saturated.

\section{Evaluation}\label{sec:evaluation}

We conducted 3 rounds of experiments to test whether technical improvement in inferring human's desired route has any effect, whether the human prefers a more natural system in expressing intent over technical efficiency, and what effect the combination of both types of systems may have.

\subsection{Experiments Setup and Methodology}

Both the starting point and the goal to take the object to are preset and are the same in all experiments. All of them are carried out indoors on a stage with OptiTrack on the ceiling to track both agents and thus know the covered distance and the duration of each experiment. The robot used is IVO~\cite{IVO_HRI2022}, which has a force sensor on each wrist and can perceive obstacles by means of front LiDAR and rear LaserScan. The experiments performed last a minimum of $36$~$s$ and a maximum of $110$~$s$.

In the first round of experiments, the usefulness of the two predictors mentioned above in achieving an effective HRI is tested. For that, each volunteer performs three experiments: one without any predictor plus one experiment with each predictor\footnote{1st round example: \url{https://youtu.be/c4aPo6WRK4M}}. The second round of experiments allows us to compare both explicit communication systems. For this, the same procedure is followed: one experiment without any communication system plus one experiment with each\footnote{2nd round example: \url{https://youtu.be/6VL41XovKJg}}.

Finally, for the third round, the predictor/communication system best rated by the volunteers in each previous round is selected and both strategies are compared. For this, each volunteer performs four experiments: a baseline without predictor or explicit communication system plus one experiment with each strategy plus one experiment with both strategies available\footnote{3rd round example: \url{https://youtu.be/mL8DQb1bK_4}}. In the three rounds of experiments, the baseline experiment is run first followed by the remaining two (three) experiments in random order to avoid statistical distortions.

At the end of each experiment, each volunteer fills out a handmade questionnaire to assess both numerically from 1 to 7 and by choosing among the different systems tested different aspects associated to an effective HRI (evaluation questions shown in the Appendix). The numerical ratings are then analyzed by means of different tests. First, the Saphiro-Wilk's and Levene's tests are applied. If the variable analyzed meets the normality condition, an ANOVA test with Bonferroni correction is applied to check whether there is a statistically significant variation ($p$\textless$0.05$), in which case, a Tukey's HSD (Honest Significant Difference) test is applied. If the normality condition is not met, a non-parametric Kruskal-Wallis test is applied followed by a Nemenyi's test if statistically significant variation is detected between the systems analyzed in each case. After filling in all the questionnaires, a short interview with three open questions is conducted, giving the volunteers the opportunity to express themselves freely: What did you think of the whole experiment? What were your feelings during each attempt? What would you improve?

\begin{figure*}[t]
    \centering
    \includegraphics[width=0.38\textwidth]{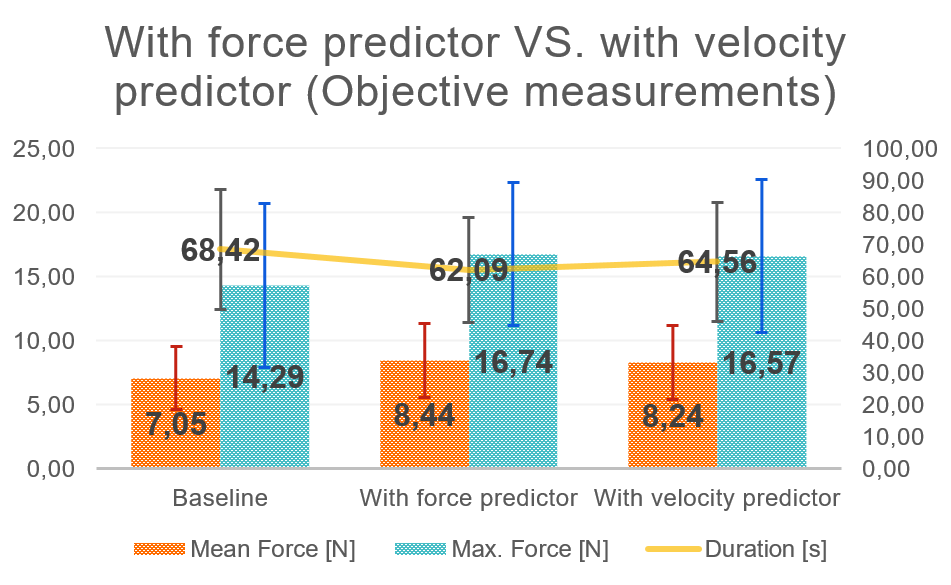}
    \includegraphics[width=0.61\textwidth]{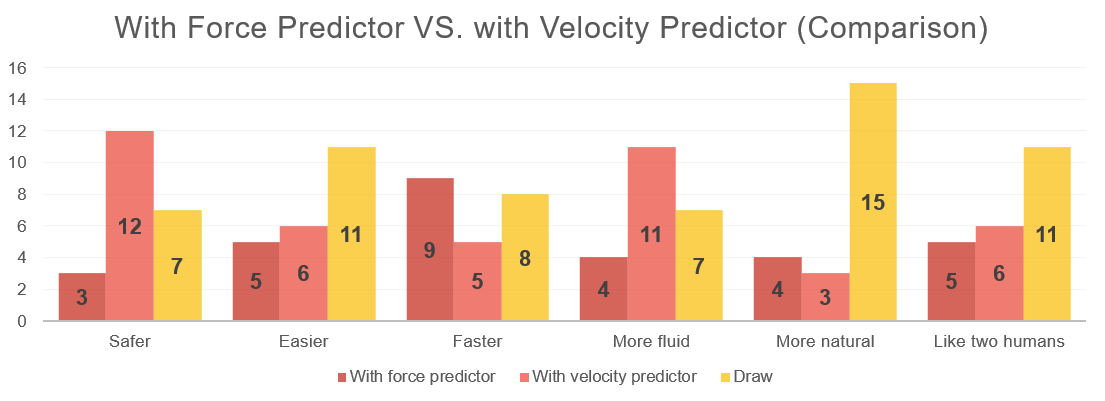}
    \caption{{\bf Assessment of objective measurements and election make by volunteers (predictor experiments).} {\it Left:} Mean force exerted in orange, maximum force exerted in light blue and duration in yellow for the three experiments. Left axis in Newtons (both forces) and right axis in seconds (duration). Bars represent std. dev. {\it Right:} Election made by the $22$ volunteers instead of valuate aspects numerically. Force predictor in dark red, velocity predictor in light red and draw in yellow.}
    \label{fig:predictors_objective_and_comparison}
\end{figure*}

\begin{figure*}[t]
    \centering
    \includegraphics[width=0.785\textwidth]{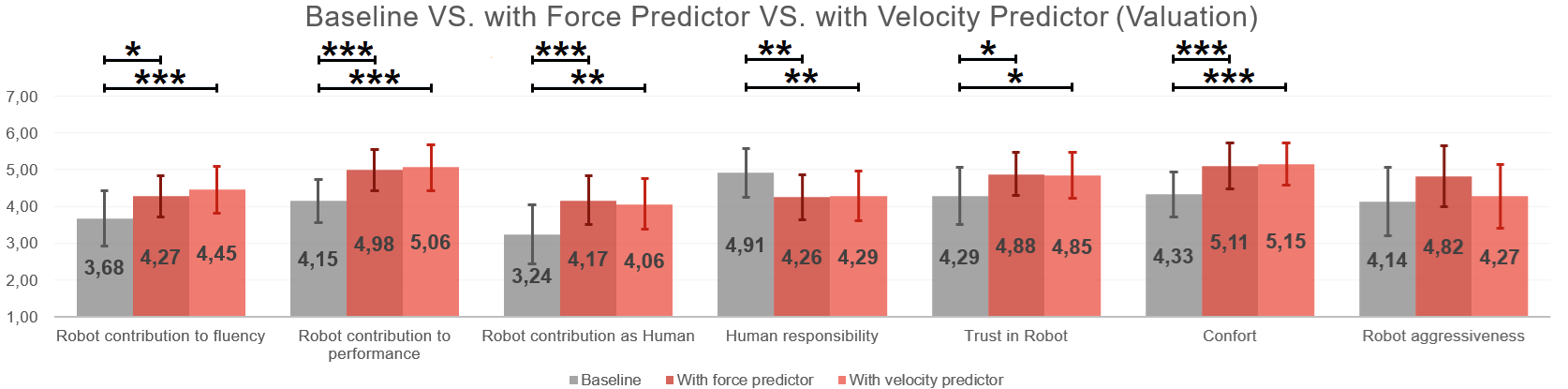}
    \includegraphics[width=0.205\textwidth]{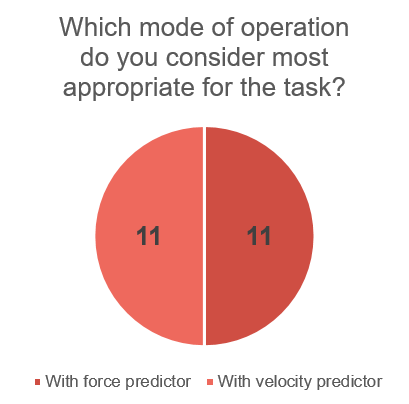}
    \caption{{\bf Assessment of the main aspects involved in the interaction (predictor experiments).} {\it Left:} Comparison among the baseline experiment (without any predictor) in gray, experiment with force predictor in dark red and with velocity predictor in light red. Valuation from 1 (very low) to 7 (very high). Statistical significance marked with *: $p<0.05$, **: $p<0.01$, ***: $p<0.001$. Bars represent std. dev. {\it Right:} Election made by the $22$ volunteers with respect to which system they prefer for the task at hand. Force predictor in dark red, velocity predictor in light red.}
    \label{fig:predictors_valuation_and_election}
\end{figure*}

\subsection{Hypotheses}

Relative to intention predictors:

{\bf H1a} - Adding a predictor to the robot's decision making system reduces the human's effort.

{\bf H1b} - Once a sufficient hit rate is reached, the human ceases to positively value technical improvements.

Relative to intention communication systems:

{\bf H2a} - Adding a way to explicitly indicate the human's intention improves multiples aspects of an effective HRI.

{\bf H2b} - The human prefers a natural communication system with a non-negligible error rate over a more robust one.

Relative to the comparison of both systems:


{\bf H3a} - A system that allows the human to directly express their intention improves multiple aspects of an effective HRI just as much as a system that attempts to infer it.


{\bf H3b} - A combination of an inference system with an explicit communication one is the option best valuated by the human and the preferred one.

\subsection{Participants}

A total of $75$ volunteers were recruited from our research institute as well as from different schools of the partner university. $22$ volunteers (age: $\mu$=$26.45$, $\sigma$=$4.02$; $23$\% female) participated in the first round of experiments performing $66$ experiments ($3$ each). $23$ volunteers (age: $\mu$=$27.36$, $\sigma$=$4.87$; $26$\% female) participated in the second performing $69$ runs (also $3$ each). Finally, $30$ volunteers (age: $\mu$=$28.32$, $\sigma$=$5.12$; $27$\% female) participated in the third round by performing $120$ experiments ($4$ each).

All the experiments reported in this article have been performed after the approval of the ethics committee of the Universitat Politècnica de Catalunya (UPC) in accordance with all the relevant guidelines and regulations (ID: 2023.05) and all the volunteers have signed an informed consent form. No volunteers were paid for participating in this study, ensuring that there is no conflict of interest.

\section{Results}\label{sec:results}

Before analyzing each round of experiments, we perform a post-hoc statistical power test to know what values we can be statistically sure of. Thus, using the criterion of $p<0.05$, for the first round ($22$~volunteers) we can detect effect sizes as low as $\eta^{2}$=$0.138$ with a statistical power of 80\%. For the second ($23$~volunteers), as $\eta^{2}$=$0.133$; and for the third round ($30$~volunteers), as $\eta^{2}$=$0.089$. All variables analyzed by variance tests are normally distributed according to the Shapiro-Wilk's test unless otherwise indicated.

\subsection{Force Predictor VS. Velocity Predictor}

To test hypothesis {\bf H1a}, we performed three objective measures (experiment duration, mean force, and maximum force exerted by the human during the experiment) in the three experiments comprising this first round. Fig.~\ref{fig:predictors_objective_and_comparison}~-~{\it Left} shows the result. No statistically significant variation is observed in any of the measures ($p$=$0.48$, $p$=$0.15$ and $p$=$0.18$ respectively) so it cannot be claimed that adding a predictor reduces the human's effort in any way. {\bf H1a} is {\bf rejected}.


\begin{figure*}[t]
    \centering
    \includegraphics[width=0.95\textwidth]{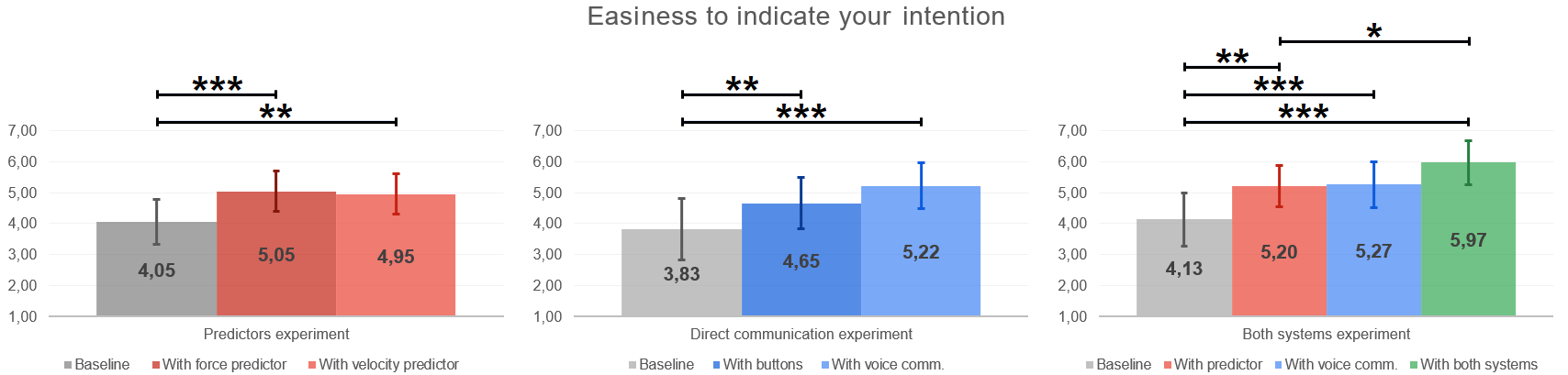}
    \caption{{\bf Assessment of the subjective easiness perceived by the user to indicate their intention.} {\it Left:} Comparison among the three experiments performed in the first round: baseline, force predictor and velocity predictor. {\it Middle:} Comparison among the three experiments performed in the second round: baseline, with buttons and with voice commands recognition. {\it Right:} Comparison among the four experiments performed in the third round: baseline, velocity predictor, voice commands recognition and both systems. Valuation from 1 (very low) to 7 (very high). Statistical significance marked with *: $p<0.05$, **: $p<0.01$, ***: $p<0.001$. Bars represent std. dev.}
    \label{fig:easiness_intention}
\end{figure*}

To test {\bf H1b}, several tests are performed. First, the volunteers are asked to rate from $1$ to $7$ multiple aspects corresponding to an effective HRI. Fig.~\ref{fig:predictors_valuation_and_election}~-~{\it Left} shows the result. As can be seen, statistically significant increases occur for all variables with these increases being similar for both predictors: "Robot contribution to performance" ($F(2,63)$=$15.89$, $p$\textless$0.001$, $\eta^{2}$=$0.335$), "Human responsibility" ($F(2,63)$=$6.88$, $p$=$0.002$, $\eta^{2}$=$0.179$), "Trust in Robot" ($F(2,63)$=$5.46$, $p$=$0.006$, $\eta^{2}$=$0.147$), "Comfort" ($F(2,63)$=$12.91$, $p$\textless$0.001$, $\eta^{2}$=$0.291$). The exceptions are "Robot contribution to fluency" where the velocity predictor generates a greater increase ($F(2,63)$=$8.44$, $p$\textless$0.001$, $\eta^{2}$=$0.211$; with force predictor: $p$=$0.011$; with velocity predictor: $p$\textless$0.001$) and "Robot contribution as Human" where the force predictor is the one with the biggest increase ($F(2,63)$=$10.74$, $p$\textless$0.001$, $\eta^{2}$=$0.254$; with force predictor: $p$\textless$0.001$; with velocity predictor: $p$=$0.0012$). No notably more positive valuations are therefore detected for the second predictor despite reducing the mean error in trajectory estimation by $30.6$\% ($0.138$~$m$ versus $0.199$~$m$) except if the subjective aggressiveness of the robot's movements (one of the questions associated to the "Robot contribution to fluency" block in our questionnaire, see the Appendix) is independently analyzed in which there is an increase in the case of using the first predictor although without being statistically significant ($p$=$0.051$).

Additionally, volunteers are asked to explicitly choose between the two predictors, accepting the draw also as a valid option. Fig.~\ref{fig:predictors_objective_and_comparison}~-~{\it Right} shows the result. In general, the velocity predictor is considered to be safer and more fluid, and the force predictor is considered to execute the task faster. The draw is the predominant choice as to which one is easier to use or which one makes the robot behave more naturally.

Finally, volunteers are asked to choose which predictor they find most appropriate for performing the task at hand, not giving the draw as an option. Fig.~\ref{fig:predictors_valuation_and_election}~-~{\it Right} shows a complete draw on this question. Therefore, {\bf H1b} is {\bf confirmed}, as the volunteers have not indicated a preference for the velocity predictor over the force predictor despite being technically superior. Some of the volunteers' comments in the post-experiments interview confirms this result. Volunteer~1.6 commented "{\it If you have changed anything between the second} (velocity) {\it and the third} (force) {\it experiment, I haven't noticed it}". Volunteer~1.13 commented "{\it The last one} (velocity) {\it seemed smoother to me but both work correctly}". 

At the end of the questionnaire that volunteers fill out after each run, we added a task-specific control question to check that they understood that the various methods they were testing were designed to allow them to indicate their intention to the robot. Fig.~\ref{fig:easiness_intention}~-~{\it Left} shows the volunteers' subjective ratings of how easy they found it to indicate their intention to the robot in this first round of experiments. There is a statistically significant increase when using any predictor but this increase is no greater when using the most objectively accurate predictor (performing a Kruskal-Wallis test since it does not pass the Shapiro-Wilk's test followed by a Nemenyi test: $H$=$17.06$, $p$\textless$0.001$; with force predictor: $p$\textless$0.001$; with velocity predictor: $p$=$0.0021$) reaffirming the hypothesis {\bf H1b}.

\subsection{Buttons VS. Voice Commands Recognition}

\begin{figure*}[t]
    \centering
    \includegraphics[width=0.38\textwidth]{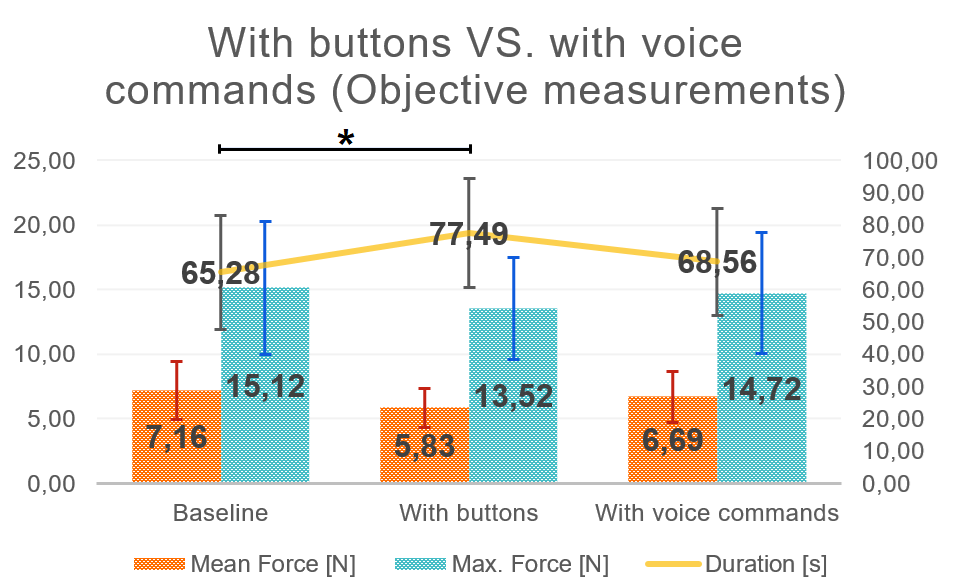}
    \includegraphics[width=0.61\textwidth]{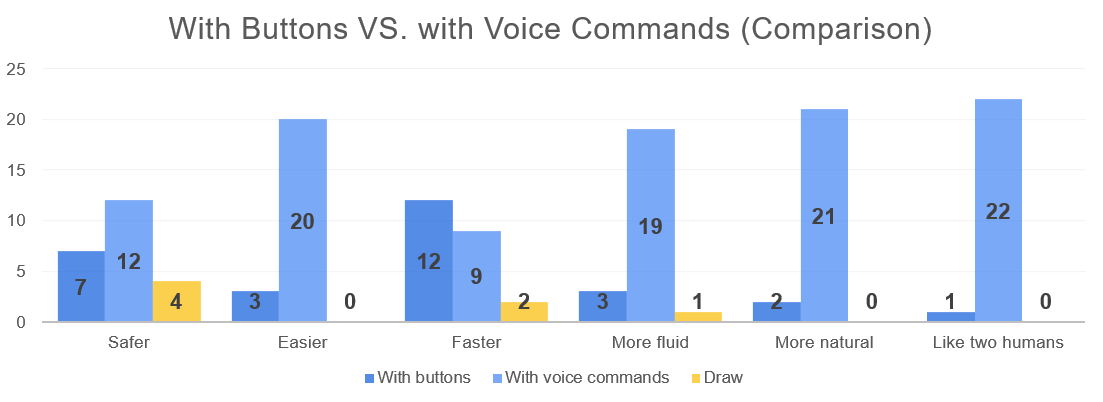}
    \caption{{\bf Assessment of objective measurements and election make by volunteers (direct communication experiments).} {\it Left:} Mean force exerted in orange, maximum force exerted in light blue and duration in yellow for the three experiments. Left axis in Newtons (both forces) and right axis in seconds (duration). Statistical significance marked with *: $p<0.05$. Bars represent std. dev. {\it Right:} Election made by the $23$ volunteers instead of valuate aspects numerically. Buttons in dark blue, voice commands in light blue and draw in yellow.}
    \label{fig:communication_objective_and_comparison}
\end{figure*}

\begin{figure*}[t]
    \centering
    \includegraphics[width=0.785\textwidth]{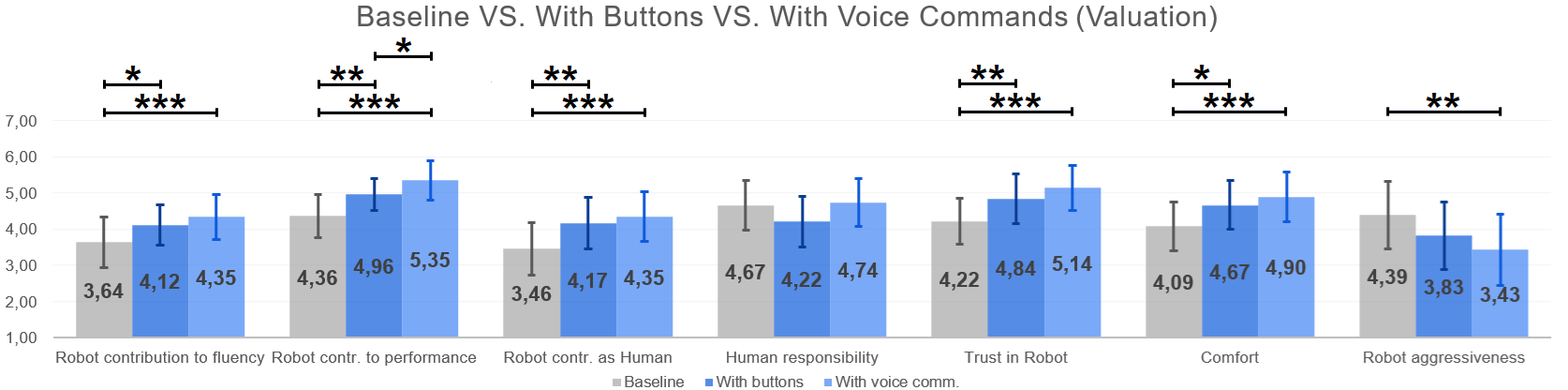}
    \includegraphics[width=0.205\textwidth]{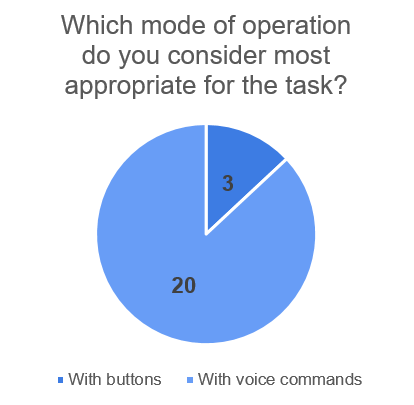}
    \caption{{\bf Assessment of the main aspects involved in the interaction (direct communication experiments).} {\it Left:} Comparison among the baseline experiment (without any communication system) in gray, experiment with buttons in dark blue and with voice commands in light blue. Valuation from 1 (very low) to 7 (very high). Statistical significance marked with *: $p<0.05$, **: $p<0.01$, ***: $p<0.001$. Bars represent std. dev. {\it Right:} Election made by the $23$ volunteers with respect to which system they prefer for the task at hand. Buttons in dark blue, voice commands in light blue.}
    \label{fig:communication_valuation_and_election}
\end{figure*}

If we perform the same objective measurements that were performed in the first round of experiments, we observe a reduction in the mean and maximum force exerted by the human in the case of using buttons although without being statistically significant (see Fig.~\ref{fig:communication_objective_and_comparison}~-~{\it Left}). This generates an increase in the duration of the experiment that we cannot consider as statistically significant as it lacks sufficient statistical power($F(2,66)$=$3.15$, $p$=$0.049$ $\eta^{2}$=$0.087$). It cannot, therefore, be indicated that there is a reduction in human effort.

To test {\bf H2a}, we use the numerical assessment made by the volunteers using the same previous questionnaire (see Fig.~\ref{fig:communication_valuation_and_election}~-~{\it Left}). A statistically significant improvement is observed in all the aspects analyzed, except in "Human responsibility" in which we lack sufficient statistical power ($F(2,66)$=$3.99$, $p$=$0.023$ $\eta^{2}$=$0.108$), being this always higher in the case of the use of voice commands. We highlight "Robot contribution to fluency" ($F(2,66)$=$7.66$, $p$=$0.001$, $\eta^{2}$=$0.188$; with buttons: $p$=$0.032$; with voice commands: $p$\textless$0.001$) and "Comfort" ($F(2,66)$=$8.67$, $p$\textless$0.001$, $\eta^{2}$=$0.208$; with buttons: $p$=$0.014$; with voice commands: $p$\textless$0.001$). There is also a statistically significant increase in "Robot contribution to performance" ($F(2,66)$=$19.63$, $p$\textless$0.001$, $\eta^{2}$=$0.373$) using voice commands relative to using buttons ($p$=$0.042$) and a statistically significant reduction in "Robot aggressiveness" (performing a Kruskal-Wallis test since it does not pass the Shapiro-Wilk's test followed by a Nemenyi test: $H$=$9.59$, $p$=$0.005$; with voice commands: $p$=$0.003$). {\bf H2a} is therefore {\bf confirmed}. For the sake of completeness, the rest of results are as follows: "Robot contribution as Human" ($F(2,66)$=$10.05$, $p$\textless$0.001$, $\eta^{2}$=$0.233$), "Trust in Robot" ($F(2,66)$=$7.66$, $p$=$0.001$, $\eta^{2}$=$0.188$). 

\begin{figure*}[t]
    \centering
    \includegraphics[width=0.38\textwidth]{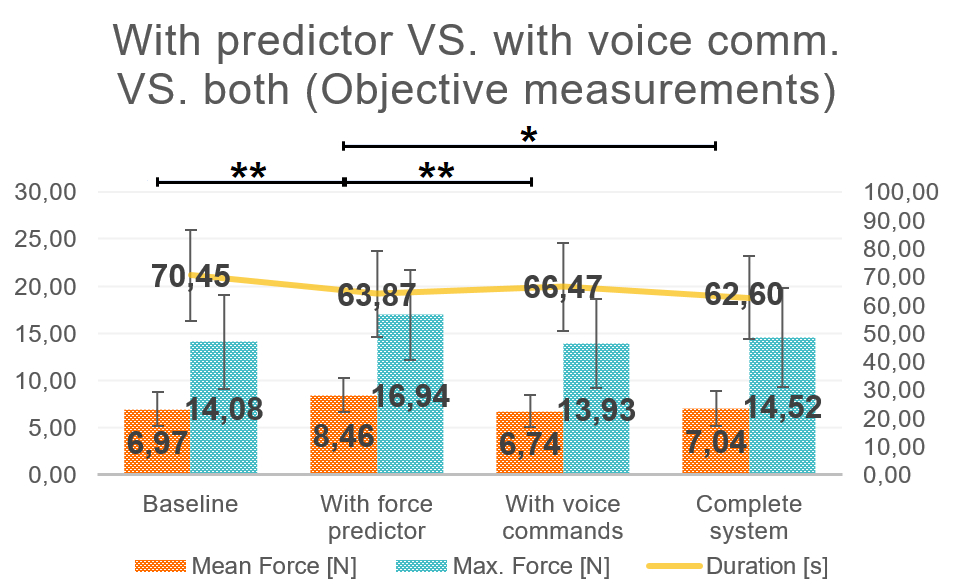}
    \includegraphics[width=0.61\textwidth]{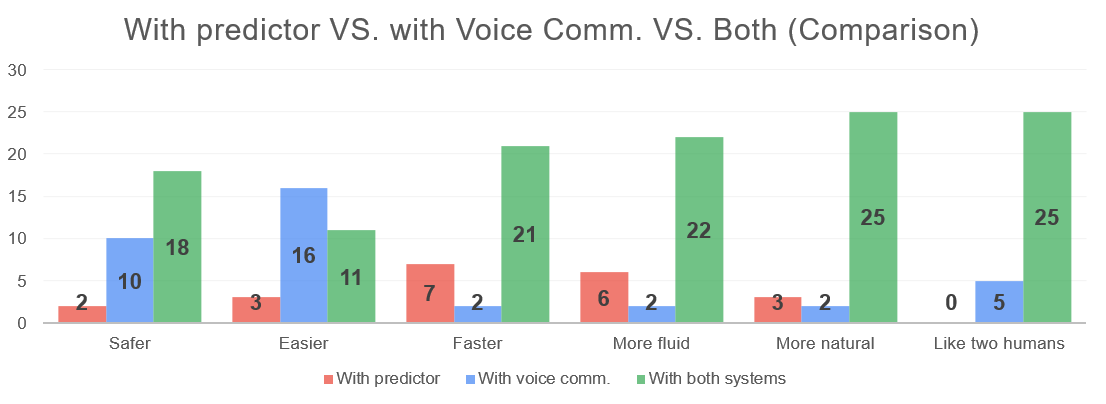}
    \caption{{\bf Assessment of objective measurements and election make by volunteers (both systems experiments).} {\it Left:} Mean force exerted in orange, maximum force exerted in light blue and duration in yellow for the four experiments. Left axis in Newtons (both forces) and right axis in seconds (duration). Statistical significance marked with *: $p<0.05$, **: $p<0.01$. Bars represent std. dev. {\it Right:} Election made by the $30$ volunteers instead of valuate aspects numerically. Velocity predictor in light red, voice commands in light blue and both systems in green.}
    \label{fig:complete_system_objective_and_comparison}
\end{figure*}

\begin{figure*}[t]
    \centering
    \includegraphics[width=0.785\textwidth]{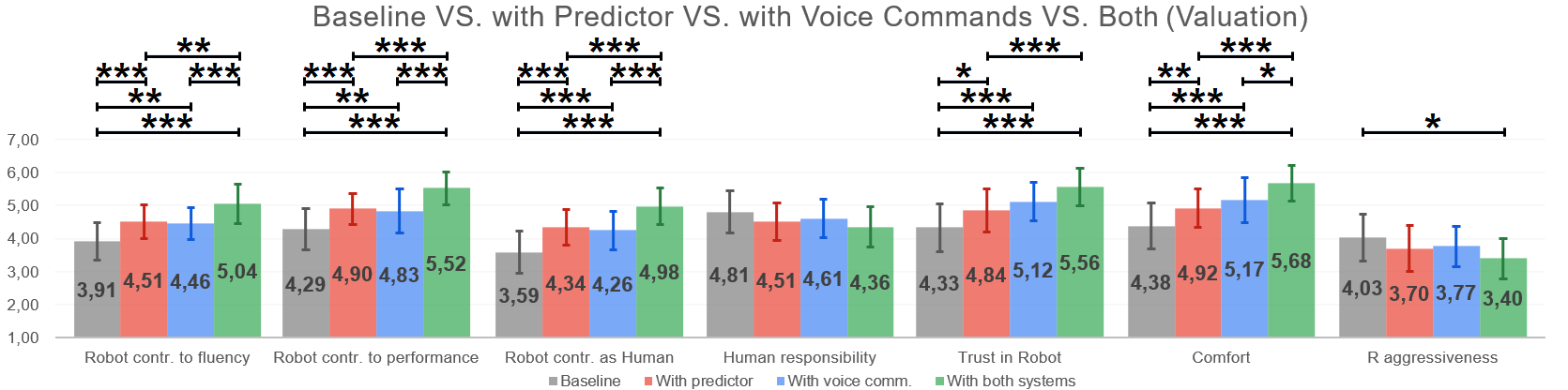}
    \includegraphics[width=0.205\textwidth]{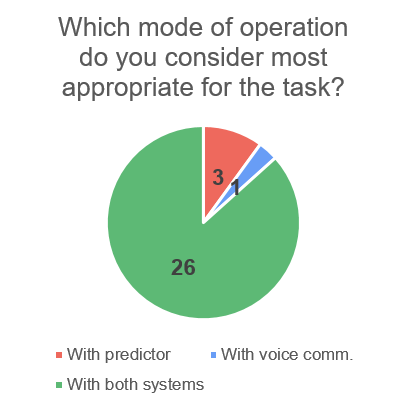}
    \caption{{\bf Assessment of the main aspects involved in the interaction (both systems experiments).} {\it Left:} Comparison among the baseline experiment (without predictor nor voice commands) in gray, experiment with velocity predictor in light red, with voice commands in light blue and with both in green. Valuation from 1 (very low) to 7 (very high). Statistical significance marked with *: $p<0.05$, **: $p<0.01$, ***: $p<0.001$. Bars represent std. dev. {\it Right:} Election made by the $30$ volunteers with respect to which system they prefer for the task at hand. Velocity predictor in light red, voice commands in light blue and with both in green.}
    \label{fig:complete_system_valuation_and_election}
\end{figure*}

To test {\bf H2b}, we asked the volunteers to choose between the two explicit communication systems (see Fig.~\ref{fig:communication_objective_and_comparison}~-~{\it Right}) with the system with buttons being considered faster when communicating and the system with command recognition coming out victorious in all other aspects analyzed. Finally, the volunteers are asked to choose which system seems more appropriate for the task being the system with voice commands chosen by 86.9\% of them (see Fig.~\ref{fig:communication_valuation_and_election}~-~{\it Right}). {\bf H2b} is therefore {\bf confirmed}.

Some of the volunteers' comments shed light on these results. Volunteer 2.8 commented "{\it Even though I sometimes have to repeat the command, I prefer to be able to talk to the robot}". Volunteer 2.17 said "{\it There is a delay until my command takes effect that with the first one} (buttons) {\it it doesn't happen but the second one} (voice commands) {\it allows me to focus more on exerting force}".

Analyzing the result of the control question in which volunteers rate the ease with which they can communicate their intention to the robot, Fig.~\ref{fig:easiness_intention}~-~{\it Middle} shows how volunteers consider that command recognition allows them to indicate their intention more easily (with a Kruskal-Wallis test since it does not pass the Shapiro-Wilk's test followed by a Nemenyi test: $H$=$19.46$, $p$\textless$0.001$; with buttons: $p$=$0.0048$; with voice commands: $p$\textless$0.001$) reaffirming the hypothesis {\bf H2b}.

\subsection{Complete System}

We take the velocity predictor (simply because it seems to generate less aggressive movements since there is not any noticeable preference between them) and the voice commands recognition system as the two preferred systems by the human for inference and direct communication respectively. With these, we perform a last round of experiments to compare both of them and their combination.

Looking at the same objective measures used above, only in the mean force exerted by the human is observed a statistically significant variation ($F(3,116)$=$5.71$, $p$=$0.0012$ $\eta^{2}$=$0.129$) (see Fig.~\ref{fig:complete_system_objective_and_comparison}~-~{\it Left}). Specifically, there is a statistically significant increase when using the velocity predictor relative to the baseline ($p$=$0.009$) and statistically significant decreases when using the voice commands or the complete system relative to using the predictor ($p$=$0.002$ and $p$=$0.014$) but not relative to the baseline ($p$=$0.963$ and $p$=$0.998$).

To check {\bf H3a} the numerical results obtained in the previous rounds cannot be directly compared as they were performed by different volunteers but require the same people to use both systems. Using the same questionnaire used in previous rounds (see Fig.~\ref{fig:complete_system_valuation_and_election}~-~{\it Left}), it can be observed that both systems produce statistically significant increases in all analyzed parameters except for "Human responsibility" where again we do not have enough statistical power ($F(3,116)$=$2.95$, $p$=$0.036$ $\eta^{2}$=$0.071$). The system with velocity predictor achieves larger increases in "Robot contribution to fluency" ($F(3,116)$=$21.42$, $p$\textless$0.001$ $\eta^{2}$=$0.356$; with predictor: $p$\textless$0.001$; with voice commands: $p$=$0.0011$) and "Robot contribution to performance" ($F(3,116)$=$22.89$, $p$\textless$0.001$ $\eta^{2}$=$0.372$; with predictor: $p$\textless$0.001$; with voice commands: $p$=$0.0022$), while the system with voice commands outperforms it in "Trust in Robot" ($F(3,116)$=$18.88$, $p$\textless$0.001$ $\eta^{2}$=$0.328$; with predictor: $p$=$0.014$; with voice commands: $p$\textless$0.001$) and "Comfort" ($F(3,116)$=$22.00$, $p$\textless$0.001$ $\eta^{2}$=$0.362$; with predictor: $p$=$0.0061$; with voice commands: $p$\textless$0.001$). {\bf H3a} is therefore {\bf confirmed}. For the sake of completeness: "Robot contribution as Human" ($F(3,116)$=$28.13$, $p$\textless$0.001$ $\eta^{2}$=$0.421$).

To verify {\bf H3b}, Fig.~\ref{fig:complete_system_valuation_and_election}~-~{\it Left} indeed shows that the system that makes use of both options is the one that scores better in all the aspects analyzed, being the only one that manages to reduce the perceived aggressiveness in the robot's movements in a statistically significant way (performing a Kruskal-Wallis test since it does not pass the Shapiro-Wilk's test followed by a Nemenyi test: $H$=$9.16$, $p$=$0.011$; complete system: $p$=$0.014$). If we ask the volunteers to choose between the three systems (see Fig.~\ref{fig:complete_system_objective_and_comparison}~-~{\it Right}), they choose the system with voice commands as the easiest to use and the complete system that makes use of both options in all other aspects analyzed. Finally, when the volunteers were asked to choose which system they considered most appropriate for the task, 86.7\% of them opted for the complete system. Therefore, {\bf H3b} is {\bf confirmed}.

Comments from some volunteers confirm these results. Volunteer 3.19 said, "{\it The predictor makes it more fluid, but being able to talk to the robot gives me extra security and peace of mind}". Volunteer 3.24 commented, "{\it They are very different approaches that I think can serve different purposes [...] give me both and I choose when and how to use each}".

Finally, in terms of how easily volunteers feel they can communicate their intention to the robot, Fig.~\ref{fig:easiness_intention}~-~{\it Right} shows once again the volunteers' preference for the system that combines both methods as they can use whichever one they feel more comfortable with at any given time (with a Kruskal-Wallis test since it does not pass the Shapiro-Wilk's test followed by a Nemenyi test: $H$=$46.39$, $p$\textless$0.001$; with predictor: $p$=$0.0013$; with voice commands: $p$\textless$0.001$, with both: $p$\textless$0.001$) reaffirming hypothesis {\bf H3b}.

\section{General Discussion}\label{sec:discussion}

The first surprising result of this study is that none of the systems tested seem to produce a reduction in human effort, understood as the mean and maximum force exerted during the task. In the case of using any of the predictors, it seems that there is even an increase that only became statistically significant in the third round of experiments. It is worth mentioning that the volunteers were aware that a predictor was being used in the experiment that this was receiving as input the previously exerted force, although without knowing which predictor specifically was in use. This is because in the second and third round of experiments, whether using buttons or command recognition, the volunteer needed to know of their existence in order to use them, so we decided to inform the volunteers of the existence of a predictor when it was being used to make the experiments comparable to each other. It is possible that this would encourage a greater force on the part of the human to make it easier for the predictor to infer their intention, understood as the desired path. At the same time, we do not believe that the non-significant reduction in the force exerted when using the buttons is so much due to the fact that the buttons encourage less effort. Observing these experiments, humans tend to pay attention to the markings on the handle before pressing any button to make sure they do not make a mistake, which means that the force exerted during that time naturally tends to be lower.

As for the comparison between the two predictors, that a considerable technical improvement goes unnoticed in the human's subjective assessment of it, confirming the hypothesis {\bf H1b}, goes against us continuing to improve our predictor. It also seems to indicate that a perfect predictor is not necessary, but simply a good enough one. This result is not entirely unexpected, as it fits with the Pareto rule or the Law of Diminishing Returns in economics. As for the two explicit communication systems, the confirmation of the hypothesis {\bf H2b} as well as comments such as those of volunteers 2.8 and 2.17 seem to indicate that naturalness is more valued over technical aspects such as lower delay and lower failure rate. These two results combined support the idea that we should pivot the current trend of attempting to infer human's intention in the best possible way towards methods that seek to improve human-robot communication making it as humane as possible.

It is worth mentioning that this work should not be understood as being against the use of predictors. In one of the experiments performed in the third round of experiments, the Wi-Fi network used for the exchange of information between the robot and the computer running the control algorithm was saturated, causing delays in the generation of the robot's speed commands according to what its sensors were detecting at any given moment. This made the interaction with the robot complex and counter-intuitive. The inclusion of the $1$~$s$ prediction made it possible to compensate for these delays and allow the human to perform the task satisfactorily. This is an illustrative example of the usefulness of using predictors. Another would be any situation in which no direct communication could take place. This is why we do not advocate discarding the use of predictors, but rather their correct combination with explicit communication systems that are as natural as possible, thus taking advantage of the benefits of both types of systems.


This explains the title of this article. The analysis of human preferences carried out in this work leads us to consider multimodality, understood as the use of multiple communication channels of different and even redundant nature, as the preferred option for humans when interacting with a robot. This allows us not to have to worry about achieving a perfect predictor. In the end, this fits with our behavior as humans: we use our prior knowledge and experience to try to predict the behavior of our peers but, when the uncertainty is too high or we simply do not know the other person well, we choose to communicate directly to avoid misunderstandings that could harm the outcome of the interaction. It is therefore to be expected that the human expects the same behavior from the robot if we want them to be considered our partners and not mere machines.

\section{Conclusions}\label{sec:conclusions}

In this work we have conducted three rounds of experiments using a collaborative transportation task to test the following: 1) Predictor-based inference systems can improve multiple aspects of effective HRI. However, there is a sufficient performance beyond which technical improvements go unnoticed by the human using them. 2) The human prefers explicit communication methods with the robot that are natural even if this is at the cost of a higher failure rate. 3) Both systems separately can achieve the same subjective ratings by the human in multiple aspects, but it is when properly combined that they achieve the best HRI in terms of fluency, trust in the robot and comfort among others.

This study should be replicated in other tasks to confirm those results. In any case, these findings can serve as a stepping stone to encourage and justify the use and evolution of human-robot communication systems that seek greater naturalness, such as natural language or gestures, even if these systems may present errors, simply because humans are willing to accept them.

\backmatter

\bmhead{Acknowledgments}

The authors want to express their gratitude to Sergi Hern\'{a}ndez for their technical support and to all the volunteers who made this work possible.

\section*{Declarations}\label{sec:declarations}


\bmhead{Funding}

Work supported under the European project CANOPIES (H2020- ICT-2020-2-101016906), the MINECO/AEI ROCOTRANSP project (PID2019-106702RB-C21 MCIN/ AEI /10.13039/501100011033) and the JST Moonshot R \& D Grant Number JPMJMS2011-85. The first author acknowledges Spanish FPU grant with ref. FPU19/06582.

\bmhead{Competing interests}

There are none potential conflicts of interest that could bias the evaluation or results of our research.

\bmhead{Ethics approval}

All the experiments reported in this document have been performed under the approval of the ethics committee of the Universitat Politècnica de Catalunya (UPC) in accordance with all the relevant guidelines and regulations (ID: 2023.05).

\bmhead{Consent to participate}

All the volunteers who participated in this study have signed an informed consent form accepting to participate in the study.

\bmhead{Consent for publication}

All the volunteers who participated in this study have signed an informed consent form accepting to publish the anonymously obtained data.


\bigskip





\begin{appendices}

\section{Example of the questions used in the questionnaires}\label{sec:A-questionnaires}

The hand-crafted questionnaires used include a first section for demographical data (age, gender, dominant hand,...) followed by the next 7-point Likert (Strongly disagree to Strongly agree) questions grouped here by categories but randomly presented to the volunteer. * means negated question. We provide the Cronbach’s alpha scores for each category.

\begin{enumerate}
    \item Robot contribution to fluency ($\alpha=0.786$):
    \begin{itemize}
        \item The robot positively contributed to the fluency of the interaction.
        \item The robot's response speed was appropriate.
        \item The robot's movements were aggressive*.
    \end{itemize} 
    \item Robot contribution to performance ($\alpha=0.843$):
    \begin{itemize}
        \item The robot positively contributed to the team's performance.
        \item The robot proposed good solutions to complete the task. 
        \item The robot helped in completing the task.
    \end{itemize} 
    \item Robot contribution as Human ($\alpha=0.887$):
    \begin{itemize}
        \item The robot contributed equally to the human in completing the task.
        \item The human-robot team worked equitably. 
        \item The robot contributed equally to the human in the team's performance.
    \end{itemize} 
    \item Human responsibility ($\alpha=0.807$):
    \begin{itemize}
        \item I had to bear the responsibility of the task for the human-robot team to function better. 
        \item I was the most important member of the team. 
        \item The team's performance depended largely on me.
    \end{itemize} 
    \item Trust in Robot ($\alpha=0.780$):
    \begin{itemize}
        \item I trusted the robot. 
        \item The robot was trustworthy. 
        \item I trusted that the robot would do the right thing at the right time. 
    \end{itemize} 
    \item Comfort ($\alpha=0.838$):
    \begin{itemize}
        \item I felt comfortable accompanying the robot. 
        \item The interaction with the robot was pleasant. 
        \item The robot's movements made me feel uncomfortable*.
    \end{itemize} 
\end{enumerate} 

The previous questions were followed by the next 7-point Likert (Impossible to Effortless) question: "How easy was it for you to indicate your intention to the robot?".

After all the experiments in the same round, the volunteer is asked to choose which system they consider "safer" and "easier to use" as well as which system they consider makes the interaction "faster to execute", "more fluid", "more natural" and "more similar to how two humans would execute it". Finally, they are asked which system they consider "most appropriate for the task".





\end{appendices}


\bibliography{./sn-bibliography.bib}

\end{document}